\documentclass[review]{elsarticle}

\usepackage{lineno,hyperref}

\usepackage{amsfonts}
\usepackage{bm}
\usepackage{paralist}
\usepackage{booktabs}
\usepackage{color}

\journal{Journal of \LaTeX\ Templates}

\begin{document}

\begin{frontmatter}

\title{Relevant Region Prediction for Crowd Counting}

\author{Xinya Chen, Yanrui Bin,  Changxin Gao, Nong Sang\corref{mycorrespondingauthor}}
\address{Key Laboratory of Ministry of Education for Image Processing and Intelligent Control, School of Artificial Intelligence and Automation, Huazhong University of Science and Technology}
\cortext[mycorrespondingauthor]{Corresponding author}
\ead{\{hust\_cxy,yrbin,cgao,nsang\}@hust.edu.cn}

\author{Hao Tang}
\address{Department of Information Engineering and Computer Science, University of Trento}
\ead{hao.tang@unitn.it}


\begin{abstract}
Crowd counting is a concerned and challenging task in computer vision.
Existing density map based methods excessively focus on the individuals' localization which harms the crowd counting performance in highly congested scenes. In addition, the dependency between the regions of different density is also ignored.
In this paper, we propose Relevant Region Prediction (RRP) for crowd counting, which consists of the Count Map and the Region Relation-Aware Module (RRAM).
Each pixel in the count map represents the number of heads falling into the corresponding local area in the input image, which discards the detailed spatial information and forces the network pay more attention to counting rather than localizing individuals.
Based on the Graph Convolutional Network (GCN), Region Relation-Aware Module is proposed to capture and exploit the important region dependency. The module builds a fully connected directed graph between the regions of different density where each node (region) is represented by weighted global pooled feature, and GCN is learned to map this region graph to a set of relation-aware regions representations.
Experimental results on three datasets show that our method obviously outperforms other existing state-of-the-art methods.
\end{abstract}


\end{frontmatter}


\section{Introduction}

Crowd counting is the task of predicting the number of individuals appearing in specific scenes. It serves as a fundamental technique for numerous computer vision applications, such as in video surveillance, public safety, flow monitoring, traffic monitoring, and scene understanding. It is also a challenging problem due to the variations of density, scale, illumination and severe occlusion.

Recently, with the development of convolutional networks (CNNs), the performance of crowd counting algorithms has been greatly improved.
Existing approaches use a CNN to estimate the density maps which represents both the spatial position and number of individuals and consequently couples the individuals' localization and counting. Although great progress has been made, state-of-the-art density map estimation based methods still suffer from two problems.
Firstly, the density map excessively focuses on the localization for its exhaustedly utilizing the spatial information of the individuals' location. It is unreasonable to force the network to accurately localize the individuals in highly congested scenarios. The reason is that each individual occupies too few pixels to be localized, which consequently harms the performance. Furthermore, the size of the Gaussian Kernel, which is used to generate the density map, is hard to adapt the variation in head scale and significantly affect crowd counting performance. It is either too small to make pedestrian of different scales distinguishable or too large to separate the pedestrian from the background.

\begin{figure}[t]
\begin{center}
  \includegraphics[width=0.8\linewidth]{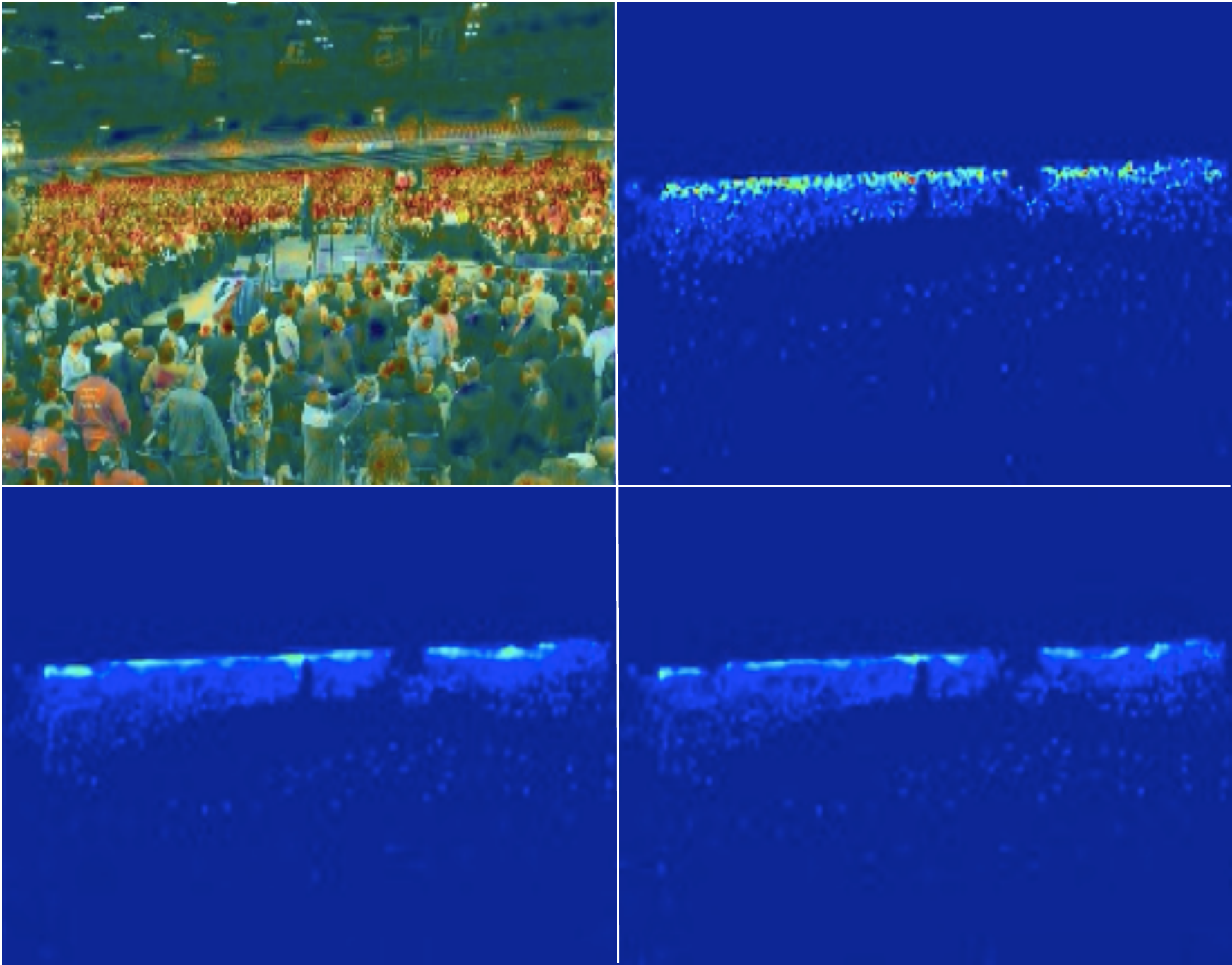}
\end{center}
   \caption{The prediction of the attentional region is refined by Region Relation-Aware Module. The first row shows the attentional region in the image and the attentional ground truth. The second row shows the attentional prediction generated with RRAM (left) and without RRAM (right). By utilizing the relation of regions, the absolute error of the attentional region decreases 14.8.}
\label{fig:region_better}
\end{figure}

 In addition, the previous approaches ignore the dependency between the regions. They adopt multiple columns or multiple regressors which major in specific regions, while regions of different density are predicted independently. Actually, the regions of different density are relevant in scenes. In congested scenes, the crowd density per square meter in the physical world is approximately constant. Due to the perspective distortion, the density changes approximate continuously along the direction away from the camera. For different views, the perspective relation varies. Moreover, the distribution of density in many scenes (such as streets, square, stadium, etc.) is governed by configurational rules. The relation can be utilized to further improve the crowd counting performance. As shown in Figure~\ref{fig:region_better}. The absolute error of the attentional region decreases by utilizing the relation of regions.

To tackle the above two problems, we propose a novel method called Relevant Region Prediction (RRP) for crowd counting, which consists of two components i.e., Count Map and Region Relation-Aware Module (RRAM). Each pixel in the count map represents the number of heads falling into the corresponding local area in the input image and the area of adjacent pixels are overlapped with each other. Thus, the network is only required to verify the presence of individuals in local area rather than accurately localize them, which forces the network to pay more attention to counting than localization. Furthermore, the system performance is robust to the area size.
The Region Relation-Aware Module (RRAM) is proposed to capture the region dependency by leveraging the power of the Graph Convolutional Network (GCN). Specifically, we represent the regions by weighted global pooled feature and build a fully connected directed graph between these regions representations to explicitly model their correlations. Then a GCN is learned to propagate information between different regions and consequently generate a set of relation-aware region representation. The weight of each edge is adjusted adaptively and thus the relationships between different regions are captured. Then these region representation are remapped to the original feature space and fused with the input feature for the more accurate prediction. Experiments show that our Count Map performs better than the density map and the Region Relation-Aware Module further improves the accuracy of the prediction.

Our contributions are three-fold:
\begin{compactitem}

\item We propose a novel labeling scheme, termed Count Map, which discards the detailed spatial information and forces the network pay more attention on counting rather than localizing individuals.
\item We design a novel region relation-aware module, which leverages the power of graph convolution network to capture and exploit the relations between regions of different density.
\item We comprehensively evaluate our model on three crowd counting benchmark datasets, and our model consistently achieves superior performance over previous state-of-the-art methods.
\end{compactitem}

\section{Related Work}

In this section, we will introduce the related work on crowd counting and graph convolutional network. 

\subsection{Crowd Counting} Various methods have been proposed for crowd counting and density estimation~\cite{Cross-scene,Cumulative,Density-aware,khan2019disam}. Early researches adopted detection based methods using a body or part-based detector to detect people and count the number~\cite{detection}. These methods are easily affected by occlusions and background clutters in highly congested scenes. To address the issues of occlusion and clutter, researchers try to deploy regression-based methods to learn a mapping from the image to the count~\cite{multiple-local-features,Feature-mining,UCSD}. Regression based methods performed well in tackling the occlusion and clutter problems. However, they ignored the spatial information due to the regression to one count.

Most recently, density map estimation is commonly used for crowd counting.  Lempitsky et al.~\cite{Lempitsky} propose to learn a linear mapping between local region features and corresponding object density maps by regression. Observed the difficulty of learning a linear mapping, Pham et al.~\cite{Pham} proposed a method which uses random forest regression to learn a non-linear mapping. After that, due to the success of deep learning, convolutional neural network(CNN) is applied for density estimation. To cope with the scale variation, Zhang et al.~\cite{MCNN} adopt multiple columns with different receptive fields by adopting different sizes of filters to adapt to variable target sizes. Sam et al.~\cite{Switching-CNN} further propose switch-CNN which choose a particular column for input patches by a density level classifier. Sindagi et al.~\cite{CP-CNN} propose CP-CNN which incorporate global context information and local context information to the multi-scale feature to generate the high-quality density map. The scale diversity is limited by the number of columns, ~\cite{SANet} increase the scale diversity by stacking the scale aggregation modules which combines filters of different sizes. Liet al.~\cite{CSRNet} demonstrate that a deeper network performs better than MCNN with a similar amount of parameters and adopt a single column architecture with dilated convolutions to deliver larger reception.  Observing that the detection based method and density estimation based method are expert in different scenes, Liuet al.~\cite{DecideNet} proposed DecideNet which adaptively choose appropriate counting method at different locations. ~\cite{IG-CNN} adopt multiple regressors which experts on the certain type of crowd to adapt with the huge diversity in images, the regressors are fine-tuned on the respective sub-dataset divided by a differential training procedure. ~\cite{D-ConvNet} learn a pool of decorrelated regressors to improve the generalization ability by managing their intrinsic diversities. ~\cite{UCF-QNRF} adopt multiple labels which include the count, density map and location, for the reason that the three statistics are related to each other. ~\cite{Rank} learns from the unlabeled data based on the prior that the sub-image contains the same number or fewer person than the super-image.

\subsection{Graph Convolutional Network} There is an increasing interest in generalizing convolutions to the graph domain, for a comprehensive review, cf.~\cite{zhou2018graph}. Advances in this direction are often categorized as spectral approaches and non-spectral approaches. Spectral approaches~\cite{bruna2013spectral} work with a spectral representation of the graphs. The convolution operation was defined in the Fourier domain by computing the eigendecomposition of the graph Laplacian. Non-spectral approaches defined convolutions directly on the graph, operating on spatially close neighbors. \cite{hamilton2017inductive} proposed the GraphSAGE which generated embeddings by sampling and aggregating features from local neighborhood nodes. Recently, GCN was explored in a wide range of area such as image classification~\cite{zhaomin}, text classification~\cite{yao2018graph}, neural machine translation~\cite{beck2018graph}. Specifically, \cite{zhaomin} builds a directed graph where each node corresponds to an object label and takes the word embeddings of nodes as input for predicting the classifier of different categories. \cite{yao2018graph} regards the documents and words as nodes and uses the Text GCN to learning embeddings of words and documents. \cite{beck2018graph} modified the syntactic dependency graph by turning the edges into additional nodes and thus edge labels can be represented as embeddings.

\section{Proposed Methods}
\begin{figure*}[h]
\begin{center}
\includegraphics[width=1\linewidth]{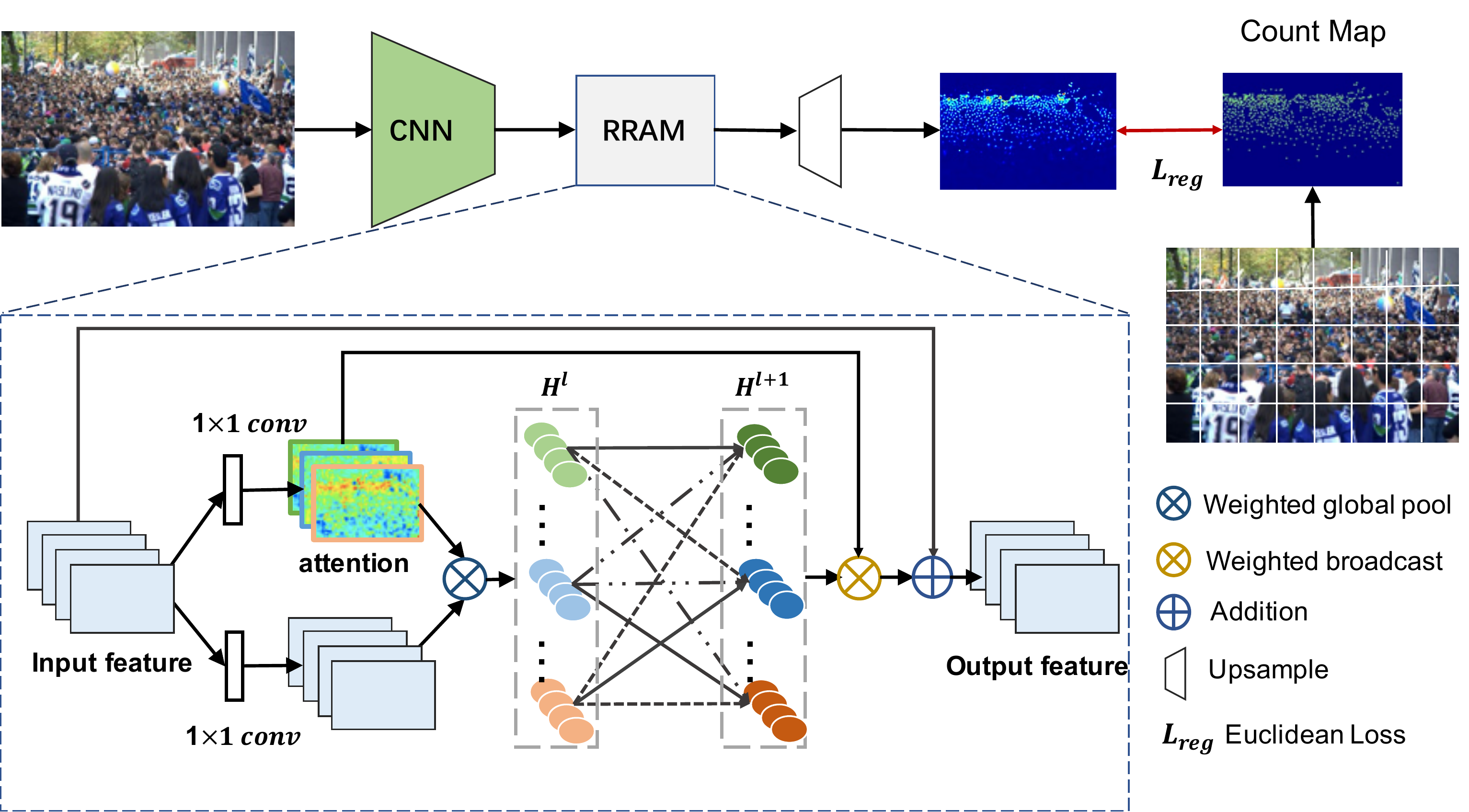}
\end{center}
   \caption{Overall framework of our proposed model for crowd counting. The input
images are fed to CNNs to obtain the appearance feature X. Then we use weighted global pooling to obtain the regions representations of different density. A directed graph is built over these regions representations to explicitly model their relationships. Based on the graph, a graph convolutional network (GCN) is learned to propagate information between regions of different density and further exploit the region dependency. In consequence, GCN generates relation-aware regions representations of different density which encode both relations and appearance information of regions. Count map is generated via applying a bilinear upsample layer on the features generated by GCN module and a $\ell_2$ loss are enforced to penalize the difference between predicted count map and ground-truth count map.}
\label{fig:pipeline}
\end{figure*}

The overall framework of our approach is shown in Figure~\ref{fig:pipeline}. The input image is fed into a convolutional neural network to extract the appearance feature. Region relation-aware module takes this feature as input and output a relation-aware feature by leveraging the power of graph convolutional network. Then the relation-aware feature is used to predict our proposed count map and a regression loss is enforced to penalize the difference between the prediction and the ground-truth.

\subsection{Count Map Labeling}

We propose a novel labeling scheme termed as Count Map to replace the commonly used density map. In standard crowd counting datasets, each training image is annotated with a set of 2D points $\{p_1, ... , p_m\}$, where $m$ is the total number of individuals. Our count map can be constructed from a location map $\bm{L}$, with $\bm{L}(p_i) = 1$ and 0 otherwise. Then a 2D sum pooling operation $SumPool2d(\cdot)$ is applied over the location map $\bm{L}$ to generate our Count map $\bm{C}$:
\begin{equation}
  \bm{C} =  SumPool2d(\bm{L};r,s)~,
  \label{equ:CM}
\end{equation}
where $r$ and $s$ is the size and stride of the pool window respectively. In all the experiments, we set $s = \frac{1}{2}r$. Naturally, when inference we calculate the integral of the count map and divide it by four as the crowd count.

To understand the advantage of our count map, we can consider taking the generation of count map to two extremes with regard to the size of the pool window $r$. The one extreme is a very large $r$. Thus, our count map is reduced to a single value which is equal to the total number of individuals $m$. At this extreme, the network is trained discarding all the spatial information provided by the individuals' location. The other extreme is $r = 1$. Here, our count map is equal to the location map which represents both spatial position and number of individuals. Especially, density map is generated by convolving the location map with a normalized Gaussian Kernel to provide a smoother training gradient. Yet, the essential idea behind the location map and density map, exhaustedly utilizing the spatial information of the individuals' location, is the same. Although the better performance of the later extreme has demonstrated the importance of the spatial information, it is still suboptimal in highly congested scenarios where each individual occupies few pixels that it is neither localized by network nor annotated by human. Thus, forcing the network to accurately localize the individuals is inappropriate and consequently harms the performance.

By choosing a proper window size $r$, our count map balances above two extremes. On the one hand, spatial information is utilized by training the network to predict the number of individuals located in different areas. On the other hand, the network is only required to verify the presence of individuals rather than accurately localize them. Experiments in Section 4.4 show that our count map outperforms both extremes by finding a balance between them.

\subsection{Region Relation-Aware Module}

The density of people, i.e., the number of people per unit area, is relevant in different regions of the image. In congested scenes, the crowd density per square meter in the physical world is approximately constant. Due to the perspective distortion, the density changes approximate continuously along the direction away from the camera. For different views, the perspective relation varies. Moreover, the distribution of density in many scenes (such as streets, square, stadium, etc.) is governed by configurational rules. Consequently, the relevant can be utilized to refine the density of one region by the regions dominating the network. Due to the success of graph convolution network to model the relationship of different nodes, we utilize it to capture the relation of density in different regions.

\subsubsection{Graph Convolutional Network Recap}

Graph Convolution Network (GCN) was introduced in~\cite{kipf2016semi} to perform semi-supervised classification on graph-structured data. The essential idea is to update the node representations by propagating information between nodes.

\begin{figure}[h]
\begin{center}
\includegraphics[width=1\linewidth,height = 0.52\linewidth]{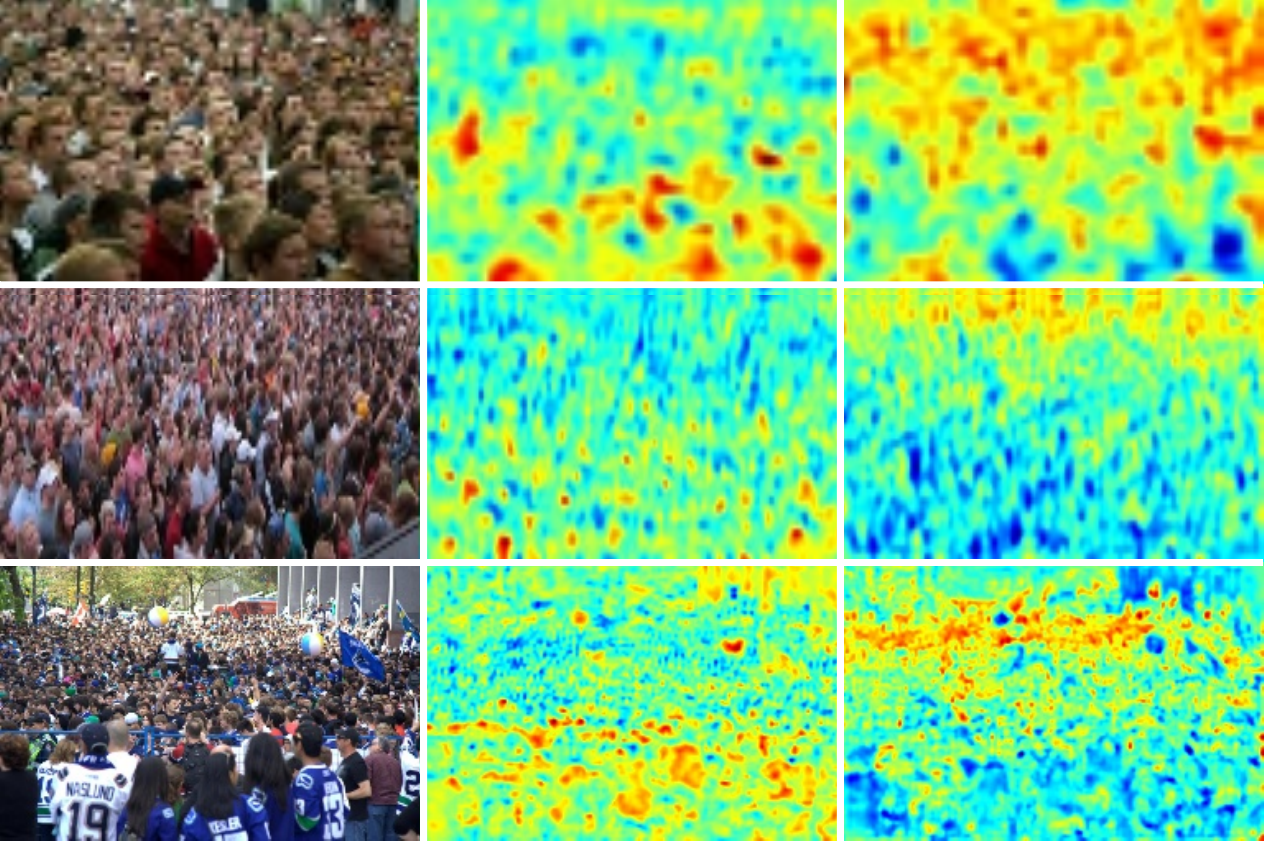}
\end{center}
   \caption{The attentional regions of two related nodes. The first column shows images in the testing set. The second column shows the corresponding node which concentrates on the region near the camera. The last column shows the corresponding node which concentrates on the region away from the camera.}
\label{fig:node}
\end{figure}

Different from the standard convolutional operations, the goal of GCN is to learn a function $f^l(\cdot)$ on a graph  $\mathcal{G}$ which takes an adjacency matrix $\bm{A} \in \mathbb{R}^{n \times n}$ and a feature description $\bm{H}_v^l \in \mathbb{R}^d$ for every node $\emph{v}$ at $l^{th}$ layer as inputs. Let $\bm{H}^l$ denote the $\emph{n} \times \emph{d}$ feature matrix obtained by stacking together all the node feature description of the graph $\mathcal{G}$. $\emph{n}$ is the number of nodes and $\emph{d}$ is the dimension of features. Then it produces a node-level output $\bm{H}_v^{l+1} \in \mathbb{R}^{d'}$ for every node $\emph{v}$. Every neural network layer can then be written as a nonlinear function:
\begin{equation}
   \bm{H}^{l+1} = f^l(\bm{H}^{l}, \bm{A})\,.
\end{equation}

Specifically, ~\cite{kipf2016semi} adopts the convolutional operations for each layer in the network $f(\cdot)$ can be represented as
\begin{equation}
\label{eq:gcn}
	\bm{H}^{l+1} = g(\bm{\widehat{A}} \bm{H}^{l} \bm{W}^{l})~,
\end{equation}
where $\bm{W}^{l} \in \mathbb{R}^{d \times d'}$ is a learned transformation and $\bm{\widehat{A}}$ is the normalized version of the adjacency matrix $\bm{A}$ of the graph, with $n \times n$ dimensions. $g(\cdot)$ denotes the nonlinear operation. In our experiments, $g(\cdot)$ is acted by ReLU and the output dimension $d'$ is always equal to the input dimension $d$.

\subsubsection{Region Relation Modeling}

The input of our Region Relation Module is a 3-D tensor $\bm{X}$ which consists of a set of feature maps and generated by the representation learning module. A $1 \times 1$ convolution $\phi(\cdot)$ is used to reduce channel dimension of the input feature maps $\bm{X}$ and then the initial density description $\bm{z}_v \in \mathbb{R}^d$ of region $v$ is generated by applying a weighted global pooling on $\phi(\bm{X})$.

\begin{equation}
  \bm{z}_v = GAP(\bm{W}_{v} \odot \phi(\bm{X}) )~,
  \label{equ:L2loss}
\end{equation}

where $GAP(\cdot)$ is global average pooling operation and $``\odot"$ represents the channel-wise Hadamard matrix product operation . $\bm{W}_{v} = \theta_v(\bm{X})$ is the attentional map for region $v$, where $\theta_v(\cdot)$ is $1 \times 1$ convolution with output channel 1. In a specific attention map, the region which has similar attributes (for example, pedestrian scale) will be activated. By learning different weights, different regions will be activated in different attention maps. After  the weighted global pooling, we obtain the features attending different regions.

Then graph convolution network is employed to model the relations between density of different regions. The input of GCN is the set of attended regions and their corresponding initial density description. For the output we want to predict a set of relation-aware region density descriptions. 

We construct a fully connected directed graph where each node represents an attended regions. Then relationships between different regions are learned by adjust the weight of each edge adaptively. Thus, the $n \times n$ adjacency matrix $\bm{A}$, representing the graph structure, is optimizable and randomly initialized. Furthermore, we add an identity matrix to $\bm{A}$, which forces each node to pay more attention to itself at the beginning of training.

Each layer $l$ of GCN takes the feature representation from previous layer $\bm{H}^l$ as input and outputs a new feature representation $\bm{H^{l+1}}$. For the first layer, the input is $\bm{Z} = \{\bm{z}_v\}_{v=1}^n$ which is generated by weighted global pooling. For the final layer, the output feature vector is $\bm{H^L}$ which has the same size of $\bm{Z}$. $L$ is the number of GCN layers.

After GCN, we use an attentional map to broadcast each node feature into a 3-D tensor and then add all generated 3-D tensors together.
\begin{equation}
  \bm{X}' = \sum\limits_{\forall v } broadcast(\bm{H}_v^L) \odot \bm{W}_v~,
  \label{equ:L2loss}
\end{equation}
where $broadcast(\cdot)$ transforms the input vector to a 3-D tensor by placing it in every position and $\bm{W}_v$ is the same attentional map used in weighted global pooling. Then we apply a $1\times1$ convolution to expand the channel dimension of feature maps $\bm{X}'$ and fuse it with the input feature maps $\bm{X}$ by addition. The module is applied between after the end of VGG.

\subsection{Learning}

The overall model consists of CNN and a RRAM module. A bilinear upsample layer is applied on the output feature maps of RRAM to generate the count map. The $\ell_2$ loss is enforced to penalize the difference between the predicted count map and the ground truth count map:
\begin{equation}
L_{reg} = \frac{1}{N}\sum\limits_{i=1}^{N}\Vert F(I_i;\Theta)-\bm{C}_i\Vert_2^2\,,
\label{equ:L2loss}
\end{equation}
where $\Theta$ refers to the set of learnable parameters. $I_i$ is the input image. $F(I_i;\Theta)$ denotes the estimated count map for image $I_i$. $\bm{C}_i$ is the corresponding ground truth count map of image $I_i$. $N$ is the number of training images. $L_{reg}$ is the regression loss between the ground truth count map and the estimated count map. To accelerate the convergence, we assist the regression loss with a cross-entropy loss which is defined as:
\begin{equation}
{L_{cls} =\sum\limits_{i=1}^{N}\sum\limits_{j=1}^{M}-\sum\limits_{k=1}^{C}y_{ijk}log(p_{ijk})~.}
  \label{equ:ce}
\end{equation} 
{Where $N$ is the number of training images. $M$ is the number of pixels in the image. $C$ is the number of category.  $p_{ijk}$ refers to the predicted probability. $y_{ijk}$ is the indicator variable. If the pixel belong to class k, $y_{ijk}=1$. Otherwise, $y_{ijk}=0$. The overall objective function is defined as:
}
\begin{equation}
   {L =  L_{reg} + L_{cls}~.}
  \label{equ:loss}
\end{equation}

\section{Experiments}

In this section, we first introduce the implementation details and evaluation metrics. Then, we report the comparison results in three popular crowd counting benchmark datasets. In the following, ablation studies and visualization analyses are presented.

\subsection{Training Details}

We adopt the modified VGG-16 network as our backbone for its strong transfer learning ability. To make the structure adapt to arbitrary resolution, we remove the three fully-connected layer. Considering the tradeoff between accuracy and resource cost, we remove the last two pooling layers. The RRAM is applied at the end of VGG followed by a bilinear upsample. Then two branches are applied for a regression task and a classification task, with (conv-3-256)-(conv-1-k) and (conv-3-256)-(conv-1-1) respectively.  (conv-kernel size-channel) denotes the convolution parameter, k denotes the number of categories.

We conduct the experiments on three public datasets. For each image in the training set, we augment it by randomly cropping 9 patches with 1/4 size of the original image, and then flipping each patch in the horizontal direction. We implement our model based on the PyTorch framework. In all the experiments,  we set the window size $r=8$ to generate the count map and use a single-layer GCN in RRAM for the better performance. The related experiments are shown in Table~\ref{table:Area size} and Table~\ref{table:gcn_ablation}. We set the batch size as 1 and employ stochastic gradient descent (SGD) as the optimizer with a fixed learning rate. To cope with the overfitting, we employ L2 regularization with the weight decay at 0.0005. The layers introduced from the VGG-16 are initialized with the weight of public-released ImageNet pre-trained model. The other layers adopt Gaussian initialization with 0.01 standard deviation. 

\subsection{Evaluation Metrics}

Following the previous works~\cite{Cross-scene,Switching-CNN,Crowdnet,MCNN}, we evaluate the performance via the mean absolute error (MAE) and mean square error (MSE) which are defined as:
\begin{equation}
  MAE = \frac{1}{N}\sum\limits_{i=1}^{N}|z_i-\hat{z_i}|~.
  \label{equ:MAE}
\end{equation}

\begin{equation}
  MSE = \sqrt{\frac{1}{N}\sum\limits_{i=1}^{N}|z_i-\hat{z_i}|^2}~,
  \label{equ:MSE}
\end{equation}
where $N$ is the number of test images, $z_i$ represents the actual number of people in the $i$-th image, and $\hat{z}_i$ represents the estimated count in the $i$-th image. The estimated count is calculated by integrating the estimated count map. Roughly speaking, MAE indicates the accuracy of the estimation, and MSE indicates the robustness of the estimation~\cite{MCNN}.

\begin{table}[h]\scriptsize
\caption{Comparison with state-of-the-art methods on ShanghaiTech~\cite{MCNN} dataset.}
\begin{center}
\begin{tabular}{l|c|c|c|c}
\toprule
 & \multicolumn{2}{|c|}{Part\_A} & \multicolumn{2}{|c}{Part\_B} \\
\hline
Method & MAE & MSE & MAE & MSE\\
\midrule
Zhang et al.~\cite{Cross-scene}  & 181.8 & 277.7 & 32.0 & 49.8  \\
Marsden et al.~\cite{Marsden}  & 126.5  & 173.5  & 23.8  & 33.1  \\
MCNN\cite{MCNN} & 110.2 & 173.2 & 26.4 & 41.3 \\
Cascaded-MTL~\cite{CMTL} & 101.3 & 152.4 & 20.0 &31.1 \\
Switching-CNN~\cite{CP-CNN} & 90.4 & 135.0 & 21.6 & 33.4\\
DecideNet~\cite{DecideNet} & - & - &20.75 & 29.42\\
SaCNN~\cite{Sacnn} & 86.8 & 139.2 & 16.2 & 25.8\\
ACSCP~\cite{ACSCP} & 75.7 & 102.7 & 17.2 & 27.4\\
CP-CNN~\cite{CP-CNN} & 73.6 & 106.4 & 20.1 & 30.1\\
IG-CNN~\cite{IG-CNN} & 72.5 & 118.2 & 13.6 & 21.1\\
Liu et al.~\cite{Rank} & 72.0 & 106.6 & 14.4 & 23.8\\
ic-CNN~\cite{ic-CNN} & 68.5 & 116.2 & 10.7 & 16.0\\
CSRNet~\cite{CSRNet} & 68.2 & 115.0 & 10.6 & 16.0\\
PSDDN +~\cite{liu2019point} & 65.9 & 112.3 & 9.1 & 14.2\\
RRP(Ours) & \textbf{63.2} & \textbf{105.7} & \textbf{9.4} & \textbf{13.9}\\
\bottomrule
\end{tabular}
\end{center}
\vspace*{-3mm}
\label{table:ShanghaiTech}
\end{table}

\subsection{ShanghaiTech Dataset}

The ShanghaiTech dataset~\cite{MCNN} is a large-scale crowd counting dataset which consists of 1198 annotated images with a total of 330,165 people. This dataset consists of two Parts: Part\_A includes 482 images in highly congested scenes with counts ranging from 33 to 3139, while Part\_B includes 716 images in relatively sparse scenes with counts ranging from 9 to 578. Following \cite{MCNN}, we use 300 images for training and 182 images for testing in Part\_A, 400 images for training and 316 images for testing in Part\_B. 

\begin{figure}[h]
\begin{center}
 \includegraphics[width=1\linewidth]{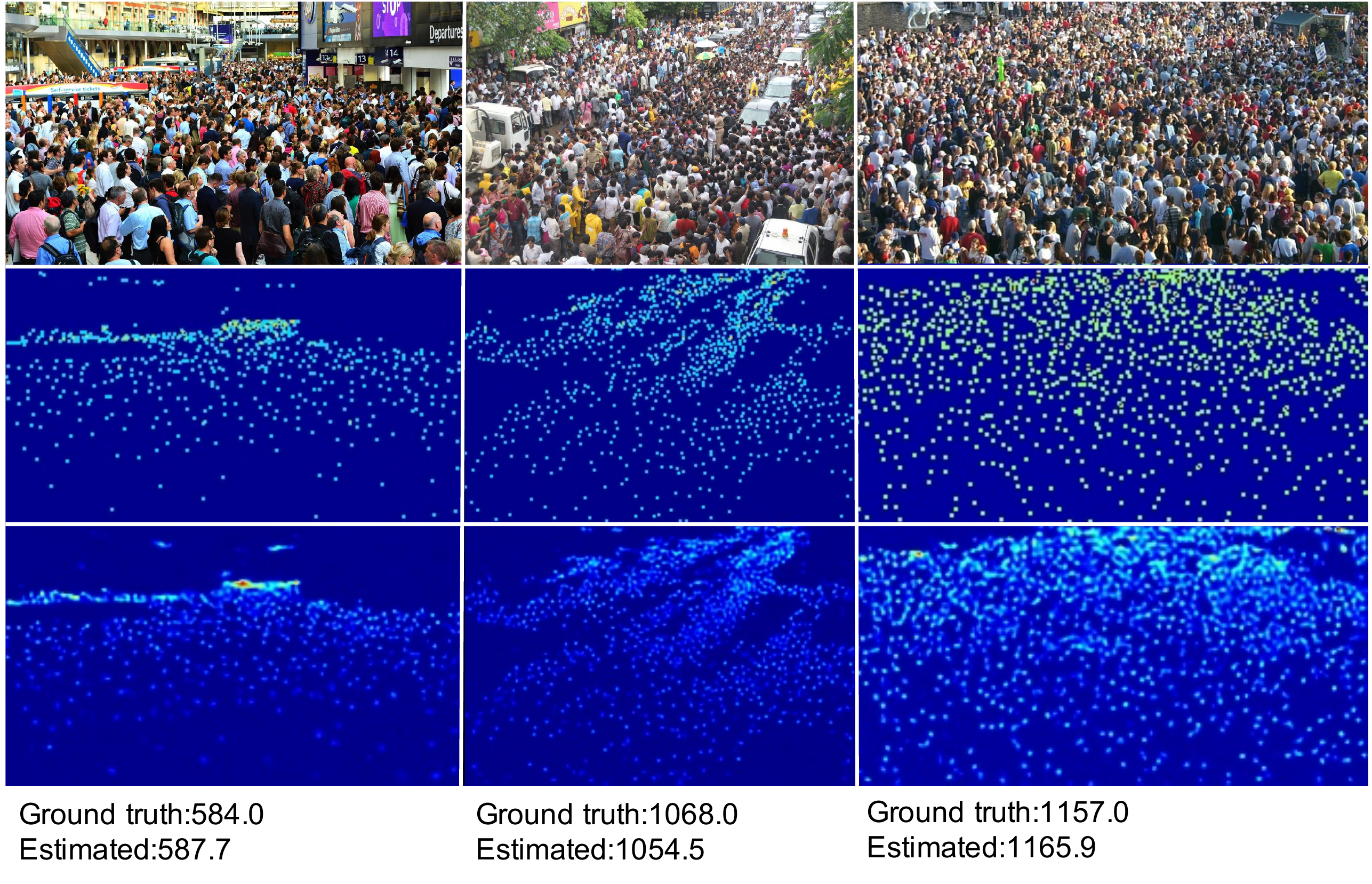}
\end{center}
\vspace*{-3mm}
   \caption{Examples on ShanghaiTech Part A~\cite{MCNN} dataset. The first row shows images in the testing set. The second row shows the corresponding ground truth. The third row shows the generated count map. The prediction is normalized together with the ground truth to obtain the heat map.}
\label{fig:ShanghaiTech}
\end{figure}

We compare our method with previous state-of-art methods on the ShanghaiTech dataset. All the detailed results are illustrated in Table~\ref{table:ShanghaiTech}. It indicates that our method achieves the lowest MAE in both Part A and Part B compared to other methods. Examples are shown in Figure~\ref{fig:ShanghaiTech}. Our model performs better than CSRNet\cite{CSRNet} which also adopts VGG-16 as backbone and applies several dilated convolutions as the backend to deliver larger reception. The better performance denotes the effectiveness of our methods. 

We visualize the attentional regions of two related nodes. As shown in Figure~\ref{fig:node}, the node which concentrates on the region near the camera is related to the node which concentrates on the region away from the camera.

\begin{figure}[h]
\begin{center}
\includegraphics[width=1\linewidth]{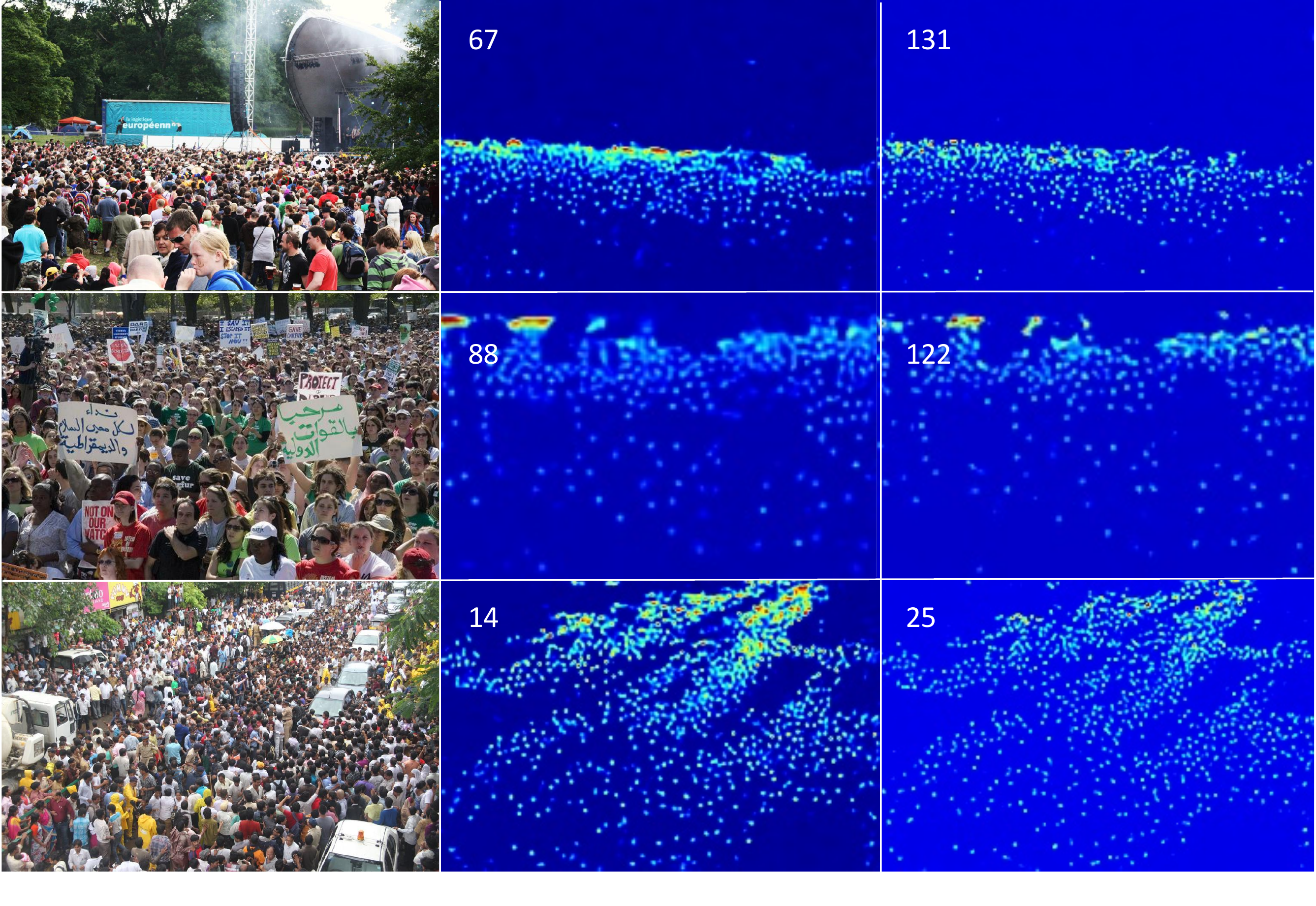}
\end{center}
\vspace*{-7mm}
   \caption{Examples on ShanghaiTech Part A~\cite{MCNN} dataset. The first column shows the images. The second column shows the prediction of count map with RRAM. The third column shows the prediction of count map without RRAM. The absolute error is shown on the picture.}
\label{fig:count_rram}
\end{figure}

\subsection{Ablation Study on ShanghaiTech Part\_A}

In this section, we conduct an ablation study on ShanghaiTech Part\_A dataset. A modified VGG-16 backbone appended a regression branch with density map output is used as our baseline which achieves 70.6 MAE and 115.2 MSE. The overall results are shown in Table~\ref{table:ablation}. Based on the baseline network, we analyze each component of our model, i.e., count map, Region Relation-Aware Module (RRAM) and classification, by comparing the MAE and MSE. We also conduct experiment on two important parameters, i.e., the area size $r$ and the number of GCN layers $L$.

\textbf{Count map.} We first evaluate the effect of our count map where each pixel represents the count in $8\times8$ area. By replacing the density map in the baseline model as our count map, we get the MAE of 65.3 and the MSE of 109.9, which is about 5.2 MAE and 5.3 MSE lower than the baseline model. The significant improvement demonstrates the effectiveness of count map. We also apply the count map to other classical methods. We implement CSRNet and MCNN and obtain better performance than the original ones. As shown in Table~\ref{table:model}, the application of the Count Map can bring improvement on both CSRNet and MCNN.

\textbf{RRAM.} To justify the contribution of the Region Relation-Aware Module, we embed it after the end of the modified VGG-16 backbone. By utilizing the correlation of regions of different densities, the predicted count is adjusted to a more accurate value. The MAE decreases from 65.3 to 63.2 and MSE decreases from 109.9 to 105.7, which validates the effectiveness of the RRAM module.
Examples of comparable prediction are shown as Figture~\ref{fig:count_rram}.

\begin{table}[h]\scriptsize
\caption{Ablation study on ShanghaiTech Part A~\cite{MCNN} dataset.}
\begin{center}
\begin{tabular}{l|c|c}
\toprule 
 Methods &  MAE & MSE\\
\midrule
Density map & 70.5 & 115.2 \\
Count map & 65.3 & 109.9 \\
Count map + RRAM  & 63.2 & 105.7\\
\bottomrule
\end{tabular}
\end{center}
\vspace*{-9mm}
\label{table:ablation}
\end{table}

\begin{table}[h]\scriptsize
\caption{Comparison with different labeling scheme on ShanghaiTech Part A~\cite{MCNN} dataset.}
\label{table:model}
\begin{center}
\begin{tabular}{c|c|c|c|c}
\toprule 
 & \multicolumn{2}{c|}{MCNN}  & \multicolumn{2}{c}{CSRNet}\\
 & Density map & Count map & Density map &Count map\\
\midrule
MAE & 99.5  & 93.2 & 67.3 & 64.1 \\
MSE & 150.8  & 145.0  & 109.2 & 101.3\\
\bottomrule
\end{tabular}
\end{center}
\vspace*{-5mm}
\end{table}

{
\textbf{Classification.} To analyze the effect of classification, we conduct ablation study on classification. The results are summarized in Table~\ref{table:Classification}. We can observe that the classification can accelerate the convergence and has minor effects on the final performance. 
}

\begin{table}[h]\scriptsize
\vspace*{-3mm}
{\caption{Ablation study of Classification on ShanghaiTech Part A~\cite{MCNN} dataset.}}
\label{table:Classification}
\begin{center}
\begin{tabular}{l|c|c|c}
\toprule 
{Methods} &  {MAE }& {MSE} & {Epoch}\\
\midrule
{Count map without Classification}  & {66.6} & {110.1} &{79} \\
{Count map with Classification} & {65.3} & {109.9} & {52} \\
\bottomrule
\end{tabular}
\end{center}
\vspace*{-5mm}
\end{table}

\begin{table}[h]\scriptsize
\caption{Comparison with different area sizes on ShanghaiTech Part A~\cite{MCNN} dataset.}
\begin{center}
\begin{tabular}{c|c|c|c|c}
\toprule
Area size & $4\times 4$ & $8\times 8$ & $16\times 16$ & $32\times 32$\\
\midrule
MAE&  67.8 &  65.3 & 66.1& 67.2 \\
MSE& 112.0  & 109.9 &107.3 & 109.2\\
\bottomrule
\end{tabular}
\end{center}
\vspace*{-4mm}
\label{table:Area size}
\end{table}

\textbf{Area size.} We conduct experiments to explore the influence of area size $r$ to the performance. We upsample (or downsample) the feature maps output by the last convolutional layer of the modified VGG-16 backbone to match the size of different count map. Results are shown in Table~\ref{table:Area size}. The count map with $r = 8$ obtains the best performance by finding a perfect balance. In comparison, a larger area size which focuses more on counting and a smaller area size which pay more attention to localization lead to worse results. Based on above observation, the hyper-parameter r can be set by gradually squeezing the range from two extremes. Note that, the count map with $r = 32$ still performs better than the density map with significantly less computation. We can observe that the MAE varies between 65.3 and 67.8 when the area size changes from $r=4$ to $r=32$, which indicates that the performace of our count map is robust to the variation in area size. 

\textbf{Number of GCN layers.} We also conduct experiments to explore the effects of different numbers of GCN layers. As shown in Table~\ref{table:gcn_ablation}, when the number of graph convolution layers increases, the crowd counting performance decreases. MAE increase $0.4$ and $1.5$ when a graph convolutional layers is added incrementally on a single GCN in RRAM module. This is probably caused by the over-smoothing problem as GCN going deeper. After GCN, the feature of each node will be the weighted sum of its own feature and the adjacent node's features, and consequently is too smoothed to be distinguishable.

\begin{table}[h]\scriptsize
\caption{Comparison with number of GCN layers on ShanghaiTech Part A~\cite{MCNN} dataset.}
\begin{center}
\begin{tabular}{c|c|c|c|c}
\toprule
Layer & 0-layer & 1-layer & 2-layer & 3-layer\\
\midrule
MAE& 65.0 & 63.2 & 63.6 & 64.7\\
MSE& 107.6  & 105.7 & 104.5  &108.9\\
\bottomrule
\end{tabular}
\vspace*{-3mm}
\end{center}
\label{table:gcn_ablation}
\end{table}

\subsection{The UCF\_CC\_50 Dataset}

The UCF\_CC\_50 dataset~\cite{Idrees} contains 50 images in extremely congested scenes. The counts range from 94 to 4543 with an average of 1280 individuals per image. It is an extremely challenging dataset due to the small dataset size, large variance in crowd count, congested scenes and large-scale change. Following the work of \cite{Idrees}, we perform 5-fold cross validation on this dataset. 

\begin{table}[h]\scriptsize
\caption{Comparison with state-of-the-art methods on UCF\_CC\_50~\cite{Idrees} dataset.}
\begin{center}
\begin{tabular}{l|c|c}
\toprule
Method & MAE & MSE\\
\midrule
Idrees et al.~\cite{Idrees} & 419.5 & 541.6\\
Zhang et al.~\cite{Cross-scene}  & 467.0 & 498.5\\
MCNN~\cite{MCNN} & 377.6 & 509.1\\
Onoro et al.~\cite{Hydra} Hydra-2s & 333.7 & 425.2\\
Onoro et al.~\cite{Hydra} Hydra-3s & 465.7 & 371.8\\
Walach et al.~\cite{Walach} & 364.4 & 341.4\\
Marsden et al.~\cite{Marsden}  & 338.6  & 424.5\\
Cascaded-MTL~\cite{CMTL} & 322.8 & 397.9\\
Switching-CNN~\cite{Switching-CNN} & 318.1 & 439.2\\
SaCNN~\cite{Sacnn} & 314.9 & 424.8\\
CP-CNN~\cite{CP-CNN} & 295.8 & 320.9\\
ACSCP~\cite{ACSCP} & 291.0 & 404.6\\
IG-CNN~\cite{IG-CNN} & 291.4 & 349.4\\
AMDCN~\cite{AMDCN} & 290.82 & - \\
Liu et al.~\cite{Rank}(Keyword) & 279.6 & 388.9\\
CSRNet~\cite{CSRNet} & 266.1 & 397.5\\
ic-CNN~\cite{ic-CNN} & 260.9 & 365.5\\
TEDnet~\cite{jiang2019crowd} &  249.5 & 354.5\\
SD-CNN~\cite{basalamah2019scale} & 235.74 & 345.6\\
\textbf{RRP(Ours)} & \textbf{216.3} & \textbf{316.6}\\
\bottomrule
\end{tabular}
\end{center}
\label{table:UCF_CC_50}
\end{table}
\setlength{\textfloatsep}{9pt}

Our method is evaluated and compared with previous state-of-art methods. 
The results are summarized in Table~\ref{table:UCF_CC_50}, it can be seen that our model significantly outperforms the state-of-the-art methods. We also conduct the ablation study on UCF\_CC\_50 dataset. Results are shown in Table~\ref{table:UCF_50_ablation}.
\begin{table} [h]\scriptsize
\caption{Ablation study on UCF\_CC\_50~\cite{Idrees} dataset.}
\label{table:UCF_50_ablation}
\vspace*{1mm}
\begin{center}
\begin{tabular}{l|c|c}
\toprule 
 Methods &  MAE & MSE\\
\midrule
Density map & 239.0  & 333.1 \\
Count map  & 228.9 &  320.9 \\
Count map + RRAM  & 216.3 & 316.6\\
\bottomrule
\end{tabular}
\end{center}
\vspace*{-7mm}
\end{table}

\subsection{The UCF-QNRF Dataset}

The UCF-QNRF dataset consists of 1535 challenging images with 1,251,642 annotations from Flickr, Web Search and Hajj footage. The training and test set consist of 1201 and 334 images, respectively. In the dataset, the median and mean counts are 425 and 815.4, respectively, and the minimum and maximum counts are 49 and 12,865, respectively, making this dataset suffering the largest crowd variation. The average image resolution is larger than other datasets, causing the absolute size of a person head to vary drastically from a few pixels to more than 1500.

\begin{table}[!t]\scriptsize
\caption{Comparison with state-of-the-art methods on UCF-QNRF~\cite{UCF-QNRF} dataset.}
\begin{center}
\begin{tabular}{l|c|c}
\toprule
Method & MAE & MSE\\
\midrule
Idrees et al.~\cite{Idrees} & 315 & 508 \\
MCNN~\cite{MCNN}  & 277 & 426 \\
Encoder-Decoder~\cite{Encoder-Decoder} & 270 & 478 \\
CMTL~\cite{CMTL} & 252  &  514 \\
Switching-CNN~\cite{Switching-CNN} & 228 & 445 \\
Resnet101~\cite{Resnet101} & 190 & 227 \\
DenseNet201~\cite{DenseNet201} & 163 & 226 \\
Idrees et al. (2018)~\cite{UCF-QNRF} & 132 & 191 \\
RAZ\_fusion~\cite{liu2019recurrent} & 116 & 195\\
TEDnet~\cite{jiang2019crowd} &  113 & 188\\
\textbf{RRP(Ours)} & \textbf{93} & \textbf{156} \\
\bottomrule
\end{tabular}
\vspace{-2mm}
\end{center}
\label{table:UCF-QNRF}
\end{table}

\begin{figure}[!t]
\begin{center}
\includegraphics[width=1\linewidth]{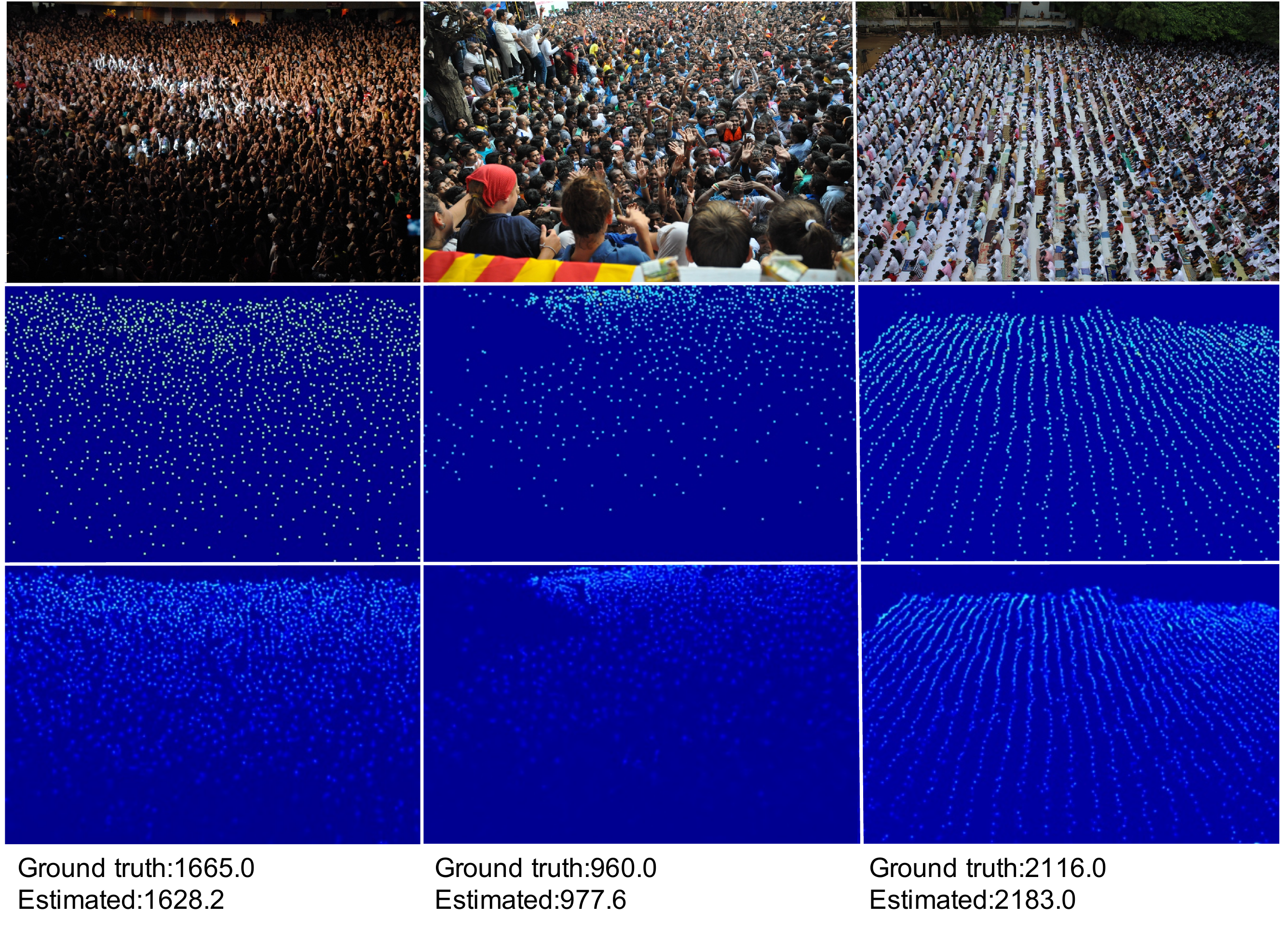}
\end{center}
\vspace*{-5mm}
   \caption{Examples on UCF-QNRF~\cite{UCF-QNRF} dataset. The first row shows the images. The second row shows the corresponding ground truth. The third row shows the generated count map. The prediction is normalized together with the ground truth to obtain the heat map.}
\label{fig:UCF-QNRF}
\end{figure}

In the whole dataset, we downsample the images to make the resolution not exceed $1080\times1920$ without changing the aspect ratio. The results of our method and previous state-of-art methods are shown in the Table~\ref{table:UCF-QNRF}. Examples are shown in Figure~\ref{fig:UCF-QNRF}. Compared to the state-of-the-art methods, our model achieves significant improvement with 25.9\% lower MAE and 18.3\% lower MSE. We also conduct the ablation study on UCF-QNRF dataset. Results are shown in Table~\ref{table:UCF_QNRF_ablation}. The ablation study of window size r is also conducted. From Table~\ref{table:Area size UCF-QNRF}. We can observe a similar trend as on ShanghaiTech\_A dataset, that a larger area size to focus more on counting or a smaller area size to pay more attention to localization leads to worse results.
\vspace*{-5mm}
\begin{table} [h]\scriptsize
\caption{Ablation study on UCF-QNRF~\cite{UCF-QNRF} dataset.}
\label{table:UCF_QNRF_ablation}
\vspace*{-2mm}
\begin{center}
\begin{tabular}{l|c|c}
\toprule 
 Methods &  MAE & MSE\\
\midrule
Density map & 111 & 182 \\
Count map  & 98 & 168 \\
Count map + RRAM  & 93 & 156\\
\bottomrule
\end{tabular}
\end{center}
\vspace*{-7mm}
\end{table}

\begin{table}[!t]\scriptsize
\caption{Comparison with different area sizes on UCF-QNRF~\cite{UCF-QNRF} dataset.}
\begin{center}
\begin{tabular}{c|c|c|c|c}
\toprule
Area size & $4\times 4$ & $8\times 8$ & $16\times 16$ & $32\times 32$\\
\midrule
MAE&  101 &  98 & 106 & 109 \\
MSE&  178 & 168 & 177 & 190 \\
\bottomrule
\end{tabular}
\end{center}
\vspace*{-5mm}
\label{table:Area size UCF-QNRF}
\end{table}

\section{Conclusions}
Existing density map based methods excessively focused on the individuals's localization which harmed the crowd counting performance in highly congested scenes. In addition, capturing the correlation between regions of different density is a crucial issue for crowd counting, which is ignored by previous methods. In this paper, we propose Relevant Region Prediction (RRP) for crowd counting, which consists of the Count Map and the Region Relation-Aware Module (RRAM). Count map is a novel labeling scheme, where each pixel represents the number of head falling into the corresponding $r \times r$ area in the input image. Thus detailed spatial information is discarded, which force the network pay more attention to counting rather than localization. Based on the Graph Convolutional Network (GCN), 
Region Relation-Aware Module (RRAM) builds fully connected directed graph between the regions of different density, where each node (region) is represented by weighted global pooled feature. Then GCN mapped the region graph to a set of relation-aware regions representation. The weight of each edge is adjusted adaptively and thus relationships between different regions is captured. Both quantitative results and qualitative visualization validate the effectiveness of the proposed method.

\section{Acknowledgment}

This work was supported by National Key R\&D Program of China (No.2018YFB1004600), and the Fundamental Research Funds for the Central Universities No.2017KFYXJJ179.


\bibliography{egbib}

\end{document}